# Certifiable Artificial Intelligence Through Data Fusion

Erik Blasch[1], Junchi Bin[2], Zheng Liu[2]

[1] MOVEJ Analytics, erik.blasch@gmail.com
[2] Univ. of British Columbia, Kelowna, BC, {zheng.liu, junchi.bin}@ubc.ca

## Abstract

This paper reviews and proposes concerns in adopting, fielding, and maintaining artificial intelligence (AI) systems. While the AI community has made rapid progress, there are challenges in certifying AI systems. Using procedures from design and operational test and evaluation, there are opportunities towards determining performance bounds to manage expectations of intended use. A notional use case is presented with image data fusion to support AI object recognition certifiability considering precision versus distance.

## 1  Introduction

Artificial intelligence (AI) and Machine Learning (ML) methods proliferate the discussion across many public and government domains. Many AI/ML principles have been presented, but the terminology typically does not include processes, metrics, and performance bounds. One example to facilitate the adoption of AI technology is that of the *Multisource AI Scorecard Table* (MAST) [1]. MAST provides the developer with a checklist for systems analysis to discern understandable systems and engender trust in AI outputs. To understand the performance of the AI, there is a need to evaluate the system for explainability (the outputs), interpretability (the processing), and accountability/attributability (decisions based on the inputs, processing, and outputs) for *certifiability*. Such a scorecard for image fusion steps enables a transparent, consistent, and meaningful understanding of AI tools [2].

With MAST, various AI/ML tools could be evaluated, such as *image fusion* [3]. Image fusion is a subset of information fusion, and is a form of data and sensor fusion. For example, in Fig. 1, both electro-optical and infrared (EO/IR) images are fed into a Deep-layer convolutional neural network (CNN). At the same time, there is both an image captioning method (e.g., neruotalk2) and the user feedback to facilitate awareness. Situation awareness from a human aligns with machine-level situation assessment [4] for coordinating information fusion, as aligned with the Data Fusion Information Fusion Group (DFIG) model [5]. As with human-machine systems, the certification process is challenging with different users and machine techniques.

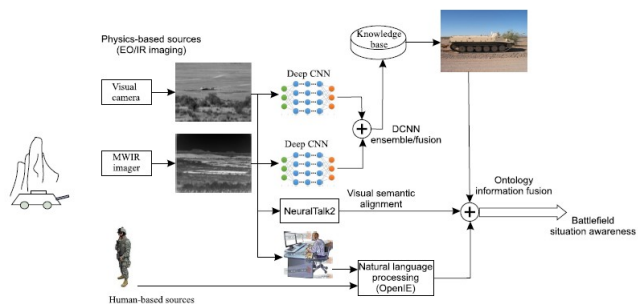

Fig. 1. Image Fusion with Deep CNN

Since certification of AI/ML systems is not standardized, there are many questions that need to be resolved for the community to adopt AI/ML technology. Key questions to the AI/ML certification process are:

1) What – type of input data;
2) Where – test location such as in a lab or field;
3) When – static assessment or run-time analysis;
4) Who – user involvement required for evaluation;
5) Which – system design or an AI/ML technique; and
6) How – metrics used to determine readiness.

Beyond certifiability, there is a need for trusted AI [6] to support scalable human-machine teaming [7]. *Scalability* can be defined by the size of parameters that can be measured, such as amount of data, entities, and routines processed. Hence, scalability can be measured by quantities. On the other hand, "trust" has many different connotations for different communities which is qualitative, requiring knowledge of user needs to be "trustworthy" [8].

Human-machine teaming (HMT) adds complexity to certification as there could be a variety of agents that include different users as agents or machine routines as agents [9]. The machine agents could forage for information to afford a user the ability to conduct sensemaking [10]. Recently, *human-agent knowledge fusion* (HAKF) [11] provides an opportunity to conduct tests and evaluations over different agents, such as seeking to certify the machine agent independently of the type of user. The challenge is how to assess systems towards certifying AI/ML machine agents, be they processes, methods, or routines.

Standards for AI are required for consistency and expectations. To seed ideas for AI standards, the MAST scorecard was proposed. MAST brings together the quantitative and the qualitative aspects of various needs for deploying AI systems within a HMT context. Relating scalability and trust is an example where the standards work towards an increasing opportunity from the machine-based AI techniques to gain scalability for the user needs. For example, within systems engineering for automation, a high-scalable approach would be to utilize the AI for its strengths in big data computations, predictive approaches to plausible futures, and fusing of different data, all of which overload most human capacities. If the AI system provides a "service", the user would likely accept it. On the other hand, if the machine-based system is autonomous, then the machine takes actions without human input. Hence, a maturity of AI adoption would include methods from which human can interact and coordinate with the AI/ML system.

HMT with AI/ML requires a maturity-model approach to adoption. Building on the MAST, the goal would be to certify AI methods within the standards of the systems-engineering pipeline. The general areas of concern would be the AI system data, training, processing, decisions, and presentation. By scoring the AI from the MAST categories, a rough qualitative estimate can be formed as to the maturity of the AI system. It is also noted the intended use would be critical to discern the appropriate deployment of the AI system.

Certifiable approaches build on performance analysis. If performance bounds are provided for the AI functions, and components, they could be certified (as in the MAST steps) and the other components gradated towards deployment. Hence, the goal would be to use MAST certifications moving from performance (e.g., uncertainty analysis - standard 2) to that of effectiveness (e.g., usability - standards 7-9).

The rest of the paper starts with Section 2 of a review of MAST. Section 3 describes the certification process and Section 4 presents certification methods such as verifiable AI, while Section 5 presents recent efforts towards certification strategies. Section 6 provides a notional example in image fusion and Section 7 provides conclusions and future directions.

## 2. Multisource AI Scorecard Table (MAST)

Over the years, there has been a continual need for systems evaluation of data fusion systems as discussed in many panel sessions [12]. While the community has yet to adopt a formal set of metrics, additional concerns abound with a lack of a set of criteria for certification. For example, information fusion supports situation awareness, but evaluating the techniques is a function of the data being used to determine awareness [13]. With the data, the measures of performance (MOPs) (e.g., accuracy) from the machine can enable measures of effectiveness (MOEs) (e.g., decision utility) for the user [14].

A different approach to certification can be through a checklist of functions that a system must satisfy, such as through a scorecard. The MAST scorecard builds on *model cards* with a rating of whether the AI techniques satisfy various criteria. Hence, if the AI/ML system has reasonably good ratings, then it can be representative of whether the system could be certified. Table 1 lists the elements of the scorecard. Key elements of certification are represented in MAST, especially the inputs (sourcing, uncertainty) as well as the outputs (consistency, accuracy, visualization). The MAST framework should be refined, but provides a notional checklist for AI/ML systems.

Table 1 – MAST Scorecard (general)

| Std | | How Determined |
|---|---|---|
| 1 | Sourcing | Properly describes quality and credibility of underlying sources, data, and methodologies |
| 2 | Uncertainty | Properly expresses and explains uncertainties associated with major analytic judgments |
| 3 | Distinguishing | Properly distinguishes between underlying intelligence information and analysts' assumptions and judgments |
| 4 | Analysis of Alternatives | Incorporates analysis of alternatives |
| 5 | Customer relevance | Demonstrates customer relevance and addresses implications |
| 6 | Logical Argumentation | Uses clear and logical argumentation |
| 7 | Consistency | Explains change to or consistency of analytic judgments |
| 8 | Accuracy | Makes accurate judgments and assessments |
| 9 | Visualization | Incorporates effective visual information where appropriate |

Since most systems would be in a dynamic setting, then robustness test is needed to certify a system for different conditions. Table 2 lists some of the criteria that can enable verification and validation (V&V) test and evaluation (T&E) criteria for an image classifier.

Table 2 – MAST Scorecard for AI Classification

| Std | Strategy of Assessment |
|---|---|
| Sourcing | Determines the data sources with the meta data suitable for training and testing |
| Uncertainty | Provides the metrics for decision making F1, probability detection/false alarm |
| Distinguishing | Distinguishes between collected data and rendered data as well as machine or human labeling |
| Analysis of Alternatives | Provides understanding of plausible target decisions for additional types for interpretability |
| Customer relevance | Accommodate customer needs |
| Logical Argumentation | Provides logical rules, statistical reasoning |
| Consistency | Achieves same results with similar data (domain adaptation) |
| Accuracy | Combines operating conditions towards condition results based on the scenario |
| Visualization | Supports explainability (heat maps) |

For the input of *sourcing*, there is a need to understand what type of data would be processed by a system that would like to be certified. Hence, the certification should come with the *model card* to characterize the expected representative data, along with the issues of the metadata.

With the careful analysis of what type of data, the system would be evaluated with that type of data for certification.

For the *uncertainty analysis*, there are many approaches to help bound what type of probabilistic (aleatoric) and epistemic (knowledge) uncertainties are included. For MOPs, such as the receiver operating characteristic curve (ROC) [15,16] would be appropriate at this stage because it measures the performance of the AI/ML system. The ROC result is not making a decision, but only outputting the results for downstream processing. The downstream MOE would require clarification of whether the right data was collected to assess completeness for decision making.

There is a need for characterizing the many types of uncertainty. The Uncertainty Representation and Reasoning evaluation framework (URREF) [17] provides an ontology which is an ongoing discussion towards mapping uncertainty types to explainability, interpretability, and transparency.

For *explainability*, the key notion is that current AI/ML systems' output results from which decisions should be analyzed from the input; however, there is a need to characterize the input uncertainty and the output uncertainty. To that end, explainability, whether complete or incomplete, requires uncertainty support towards the use of AI/ML, and/or multimodal data fusion of AI/ML systems. Hence, with the uncertainty analysis, the decision output of the AI/ML can determine whether or not the machine analytics augments user decision needs.

For *interpretability*, the rigor of the theory and mathematics of any AI/ML system can benefit from the detailed knowledge of the parameters of the models for the predictability. For example, Tomsett *et al.*, [18] looked at trust calibration through uncertainties. The uncertainty awareness provides an understanding of the model performance and the uncertainties. For object classification, the features selected with the convolutional neural network (CNN) should be mapped to the characterization of the targets. Hence, the URREF metrics associated with the comprehensive analysis of object assessment can support the certification bounds. Together, the amount, type, and association between different data sources, perspectives, and aggregation over time afford interpretability of the system - conditioned on uncertainty.

For *transparency*, the obvious analysis is a system-level understanding of the data pedigree and the reasoning strategy. The URREF transparency analysis includes data handing, data reasoning, and data reporting. Hence, data turned into information should incorporate the system contextual knowledge that supports the information fusion pipeline. Likewise, the use of the model cards [19], along with the MAST checklist, can improve transparency and fairness. Furthermore, model cards support accountable AI; in that if something goes wrong, the model card can be used to verify that the appropriate conditions for which the intended use of the AI/ML were followed.

For the outputs from the AI/ML system, MAST has a checklist for consistency, accuracy, visualization [20]. *Consistency* would make sure that the same performance was repeated. Accuracy would come from the characterization of the uncertainty. Finally, since the AI/ML system is part of a processing pipeline, then the outputs should be in a user-friendly interactive display to support decision making.

The use case in the paper is image fusion for automatic target recognition (ATR), but prior strategies for certification can be used for image data fusion adoption.

## 3. Certification Process

The certification process aligns with standard testing and evaluation (T&E) that is integrated with verification and validation (V&V) methods. Hence, there is verified AI and certifiable AI.

Fig. 2 presents a notional certification process consisting of four levels including conceptual idea, processing design, implementation test, and managed deployment. From these levels, certification is a function of the AI/ML in the environment for which it satisfies robust performance.

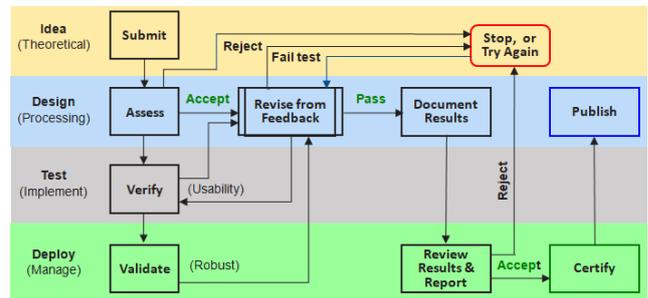

Fig. 2. Certification Process

In the *idea* phase, the conceptual method is submitted for consideration. Typically, the theoretical justification is based on some merits, such as an AI/ML method to process the available data. Inherently, the domain of application, type of data, and justification of need are available towards motivating the interest in the system. Also, it is noted that all future phases are subject to an analysis on whether the development meets the criteria in the stop/try again. Hence, even in the idea phase, there is a need to determine if the AI/ML system is worthy of interest. For example, it is likely that if the intended AI/ML is for autonomous cars, then AI/ML for scene characterization is a good idea, whereas an AI/ML for traffic light analysis could be overkill in that for a standard traffic light analysis (assuming lights are operating normally and cars oriented correctly) there are only three states, while for the scene characterization there is an unlimited set of possibilities. Once the AI/ML idea has merit on paper, the next step is to design the system.

In the *design* phase, the selection of the type of AI/ML method would be determined. For example, CNN is good for images, while Recurrent (RNN) is a choice of dynamic signals data. Clearly, most AI/ML systems are focused on the design phase as data is collected, processed, and analyzed. Initial results are presented along with the data available and analysis. If the initial design does not meet minimum standards, new methods are tried and the knowledge provides feedback to subsequent iterations. Typically, the first design is rejected and then the next iteration is developed to try again. The initial tests demonstrate some of the parameters and metrics of the AI/ML as well as the dimensionality of the problem. Hence, the testing conducted begins with simple examples and then has to consider more relevance through V&V.

In the *verification* phase, the goal is to determine if the AI/ML is performing as intended. The first verification is based on the intended use case in mind. However, since the use of AI/ML is considered for productization, then the verification should meet the requirements of the academic, industry, and government standards. As the use of the AI/ML tools requires consideration of the type of application, there are associated variations, such as whether the AI systems are for simple building security or a medical operation. For security, the AI/ML could alert a human guard on duty to go can verify the system, whereas a complex medical operation might not have the luxury as the surgeon cannot correct for the mistakes of the AI/ML deviation. Thus, verifying the AI/ML tool should be specific to domain complexity. Even then, the intended use requires documentation of findings for further review.

The last phase of *validation* requires appropriate considerations of all the operating conditions expected. For the operation conditions; it is the sensors that collect the data, the environment and variations, the intended uses, as well as objective and subjective assessment. For validation there are not only the quantitative MOPs, but also the qualitative issues to be considered, such as how the human-machine teaming would result. Clearly, if the environment is known (e.g., indoor with common lighting conditions for a manufacturing facility), then the validation could be easier. However, when there are an enormous amount of possibilities (e.g., autonomous car in all-weather conditions), then there is a rigorous set of standards to consider to determine the robustness of performance. As with the first three phases, there could be feedback and analysis for iteration. Once many of the tests are conducted and feedback in the wild, then the subsequent reviews should be scrutinized. Achieving a threshold of reliability over the performance metrics, then the system could be certified. Along with the certification comes with managing expectations of what is the intended and possible use through a published manual. The published manual should come with appropriate warnings. Hence, the MAST approaches and the AI/ML model cards provide a soup label of the data expectations, performance in certain conditions, as well as a series of information that clarifies what the AI/ML system can and cannot do.

It is to be noted that the certification process differs from regular methods of testing because during the iterations, the publication and supporting manuals must be updated based on the performance results, but also what limitations are associated with the AI/ML.

The motivation for certifiable AI also should come with *accountable AI* concerns as well as potential details that support the ethical analysis [21]. The warnings could be used to help with the AI/ML deployment such that an AI/ML tool is not used inappropriately for situations in which it was not designed for. For example, in an autonomous car/UAV, the system is dependent on the data and the types of environments where/when the system works. To help with certifiable AI, when the system engineering detects changes in the application, it could even turn off the AI/ML system. One example is that of having the autonomous car AI/ML system turn off when the imagery data has too much noise (e.g., from fog) and thus return the controls to the driver. Further analysis should be warranted to systems for dynamic design making, such as for medical applications.

## 4. Certification Methods

### 4.1 Types of Certification

The definition of certification is:

- *the action or process of providing someone or something with an official document attesting to a status or level of achievement.*,
- *to attest as being true or as represented or as meeting a standard; and/or*
- *to recognize as having met special qualifications (as of a governmental agency or professional board) within a field*

Among the many directions of AI, there are recent efforts in certifiable AI. Typically, there are certification authorities, processes, and metrics with three approaches.

The first approach is *quantitative bounds* on performance from standards methods of assessment such as for image processing, text analytics, and signals processing. For the image processing, it is the accuracy of object detection, text analytics of entity resolution, and signals processing is signal-to-noise ratios. Each of these metrics has bounds of performance that are standard for sensing and have been adopted to demonstrate the benefit of AI systems

The second method is the *qualitative appreciation* of AI to support decision making. For example, outputs of a heatmap can show where an object is recognized in an image and the user can infer that the AI system helps to manage where to look. For text analysis, it is a response to the query. For the third case, it is the signals analysis for detecting patterns and anomalies which inform the user.

The third approach is *documentation of use*. In this case, when some method or product is certified, it also comes with the manual for intended use [22].

### 4.2 Examples of Certification

Among the many types of certification that support the public domain, three examples related to AI/ML include hardware (safety), software (security), and data (processing). There are many other examples, such as those in Table 3; however, it is noted that for many systems, the human is also "certified" in many contexts, such as driver's license. For a driver's license, it is understood that the machine (hardware, software) has been certified from which the human is thus evaluated using a certified car (machine), from which the driver is then assessed. Humans are also certified based on knowledge, academic exams, and practical service (e.g., physician). Examples of user-based certifications also exist for developing software and analysis, such as the Certified Information Systems Security Professional (CISSP) [23].

Recently, there are many data analytics programs available, but they are not referenced here as there is yet to be a comprehensive community standard. Thus, in future AI/ML HMT, there could be a certification process that mirrors a driver's license. By that, beyond a data analyst certification that is either general (like a physician), or a tool (like a security analyst); the goal is to first certify the hardware and software combined with the domain (data type and application) as comparing to a driver license to a pilot's license in which both use hardware/software platforms, but use and collect different data.

Table 3: Certification Examples

| Example | Certification |
|---|---|
| Hardware - Electronics | Quality Control Rating |
| Hardware - Platform | Airworthiness |
| Sensor | Calibration specifications |
| Software - Routine | Processing time |
| Software - System | Assurance |
| Human- Novice | Drivers License |
| Human- Expert | Medical License |
| Human-Software | Security Certificate |
| Human-Data | Data Analytics Certificate |
| Human-Data Domain | Data Analysis License |

The US Department of Defense issued MIL-STD-882E for system safety in 2012 [24]. Among the guidelines include system safety, hazards, risk, and software issues towards certification. For the *system safety*, there are categories of safety-critical, -related, and -significant which align with severity measures. The eight steps include documenting that of:

1. Determine the approach (e.g., safety)
2. Identify issues (e.g., hazards)
3. Assess perform (e.g., risk)
4. Establish (e.g., risk) mitigation measures
5. Reduce (e.g., risk), increase (e.g., safety) measures
6. Verify and Validate (e.g., risk reduction)
7. Accept performance (e.g., risk)
8. Mange life-cycle (e.g., risk)

A key message is that certification comes with an acceptable *risk level*. The idea is beneficial for AI/ML systems, especially when there is not enough training data to meet all situations.

### 4.3 Certification Standards

As with the human (license), hardware (performance) and software (assurance), additional checks include the safety standard, such as the MIL-STD-882E. The challenge is in aligning existing standards, such as the safety standard to software, would not be applicable to a dynamic AI software-based system. For example, many AI routines learn on the data available and as the data changes, so does the AI capability. Hence, certifying at operational test and evaluation (OT&E) for deployment means that once deployed, the AI model should adapt. Thus, there needs to be new methods of certifying AI systems based on the data source, user's task contribution, and meaningful results.

Examples of certifiability typically include some measures of performance or proficiency. For example, for airworthiness, the Federal Aviation Administration (FAA) currently utilizes the Radio Technical Commission for Aeronautics (RTCA) DO-178C standard for certification of airborne software [25]. Within the software, agile, model-based, and experimental methods help determine whether some platform is airworthy.

One example of human-machine teaming is for air traffic controllers and pilots of remote control unmanned aerial vehicles (UAVs), as shown in Fig. 3. Using the directives from the aviation industry, then the methods could be applied to the AI/ML. Inherently, the certification of software and hardware has best practices of what types of data, procedures, and intended use are in the directives. A similar approach could be applied to AI/ML systems.

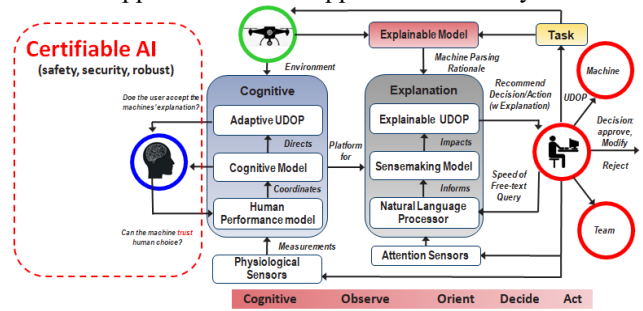

Fig. 3. Human-machine Teaming for UAVs

One example of an intended use is for UAV to be edge processors for surveillance [26,27]. Since the UAVs are at the edge, then the certification of these devices would have less stringent aviation needs than that of an aircraft carrying passengers. However, if the UAV is operating in urban airspace, other criteria are needed, such as the use of the URREF to support the coordination of the various stakeholders [28,29]. As shown in Fig. 3, the challenge is what is being certified. The first case is to certify the user who opertates a UAV through a console as a user-defined operating picture (UDOP). The UDOP could be a general type of UAV display that interacts with the machine, assuming that the air traffic controller (right side) interacts with a team and other machines. The other machines provide contextual support for weather, navigation, and airspace controls. Likewise, AI/ML can support the detection of the airspace to include the operating UAVs and aircraft [30]. Since the aviation community has rules and regulations, then the tasks are explainable with elements of the geography of the airspace, sensible flight rules, and language (e.g., notice to airman).

Another concern is the cognitive agents of the human and the UAV control. From this example, when the human interacts with the system there are human performance characteristics. Thus, certifiable AI for a UAV includes safety, security, and robustness. If the UAV is conducting surveillance, then the output of the UAV AI/ML might be image fusion that includes the quantitative performance (trust) and the qualitative perception (trustworthy).

Recently, the Society of Automotive Engineers in 2019 prepared for the use of AI/ML in automotive and aeronautical systems with the preparation of AIR6983:

*Process standard for deployment and certification/Approval of Aeronautical Safety-Related Products implementing AI* [31]. The focus is on generating guidelines for aircraft systems the utilize AI methods, including operations, environments, and avionics functions. The goal is product certification and assurances as guidelines towards design verification and requirements validation, especially in concern for safety. The document highlights best practices for the industry to follow for complying with the regulations, meetings standards, and assessing the usefulness of the guidelines in platform development. Safety is critical in aviation systems, so the standard would support aircraft, but then additions can be leveraged for examples of UAVs for emerging applications of air surveillance, package delivery, and fire monitoring.

An emerging element that relates to public safety is that of verified AI.

### 4.4 Verification AI – Theory for Certification

While there are many methods and definitions of *verified AI*, in 2020, Seshia, Sadigh, Sastry [32] provide a good guideline. The goal was to create a formal specification amongst the challenges of the unknown variables, model fidelity, and human coordination that exist with AI systems consisting of data inputs, multiple parameters, adaptable use, and varying contexts. The goal is for the verification of models and data for efficient and scalable design. They suggest five principles:

1. Use an introspective, data-driven, and probabilistic approach to model the *environment*;
2. Combine formal specifications of end-to-end behavior with hybrid Boolean-quantitative formalisms for learning system's *components*;
3. Develop new abstractions, explanations, and semantic analysis techniques for ML *processes*;
4. Create a new class of compositional, randomized, and quantitative formal methods for data generation, testing, and verification, for scalable *designs*; and
5. Develop techniques for formal inductive synthesis of AI-based systems for safe learning and run-time assurance *requirements*.

Another fascinating approach for verified AI is by Everett [33] for control by proposing how to certify the safety, performance, and robustness properties of learning machines. Given real-world uncertainties, methods of reachability, verification, and adversarial examples of NN are investigated. Robustness bounds are provided for control using a control *Q-value in a DNN*. The DQN was developed for human-level control which can provide a method of human-machine certification. Fig. 4 highlights that there is a region of robust performance and when there are adversarial uncertainties (such as from sensor noise), that might be at the edge of the certifiable region of the method selected. The hypothesis is that if there are multiple sensors, then by probing the system with a signal, fusing with other data, and using context; then the certifiable region could expand to cover adversarial situations.

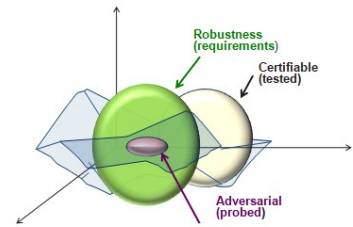

Fig. 4 Certifiable Bounds.

Another case is to use a model generator such as a *generative adversarial network* (GAN). The GAN increases robustness by generating samples that are plausible, but out of bounds of the current data available [34]. With data examples from the real world and that simulated from the GAN, a discriminator (or classifier) can reason over different situations. The classifier can be updated based on new data, data from another sensor, or the changing context. Like a classifier agent, a human agent can query for information to gather new data, whether real or simulated. Together, the system can be more robust to the changing environment.

The AI is part of a system which consists of the human, machine, AI/ML methods, and the data. Certifying the human separately from the machine; does not necessarily certify the "system." Hence, routines are needed for human trust, machine scalability, data quality, and hardware size, weight, and power (SWAP). Recent elements of certifiable AI build on the previous certification methodologies.

## 5. Certifiable AI – a Strategy

The interest in certifiable AI is currently being assessed. Basically, from Fig. 5, the AI/ML certification is after the data (as it is assumed that there are some standards for an *authoritative data* construct, see [35] for an example). The AI/ML is in a pipeline and the goal is to coordinate the T&E through V&V. Once initial tests are conducted, then further operations and maintenance can be considered for the life-cycle management certification of the system (e.g., expected types of dynamic data to include adversarial uncertainties and attacks).

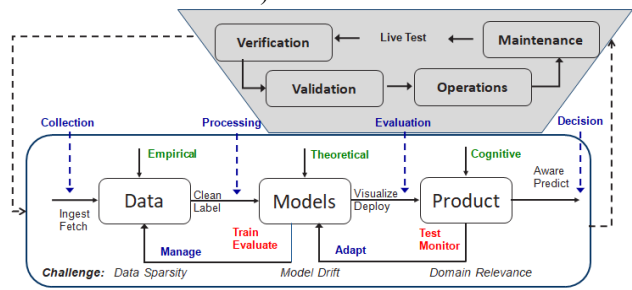

Fig. 5. Certification in an AI/ML pipeline

For certifiable AI, the issue is to determine where and what is being certified. For purposes of argument, we will assume that the data is legitimate and that the concern is for the AI/ML system. In MAST, accuracy and consistency are called out as key elements which form the basis of our approach. When the data varies, there should be accuracy and accuracy which can be evaluated for V&V and then for consistency. It is the operations and maintenance of the AI/ML system.

Certification of AI has many forms such as (1) *model certification* [36] in response to imbalanced, cost-constrained, active, transfer, and quantified learning, among others; (2) robustness performance assessment over the sensors [37], environments, and applications [38], (e.g., structural health monitoring [39,40,41]) and (3) software assurance such as blockchain [42]. Also, there is a need to consider the data, such as balancing privacy and security in certification standards [43,44]).

For imagery collected from airborne sensors, there have been ideas for methods of certification that leverage elements of camera calibration [45,46,47]. When collecting the imagery for exploitation, certification can include elements of stability and precision robustness [31], checksum on DNN layers for assurance [31], and local robustness that requires that all samples in the neighborhood of a given input are classified with the same label [48]. Many works have focused on designing defenses that increase robustness by using modified procedures for training [49]. Finally, the goal is to have fast and certifiable methods that are robust to outliers with performance guarantees [50].

Building on these ideas, we seek to generate a method for certifiable AI-based image fusion methods.

## 6. Example for Image Data Fusion

The experiment includes image fusion with contemporary AI/ML methods. The Automation Target Recognition (ATR) Algorithm Development Image Database was selected as a use case which includes (1) mid-wave infrared images (MWIR), (2) gray-scale visible image (VI), (3) and motion images (MI) generated from two consecutive visible frames/images [51]. Amongst all the methods from Zhang, *et al*. [52] and the statistics for the performance measures from Liu, *et al*. [53], various trends were observed, of which a sample is shown for three distances, three views, as well as many image fusion methods. For purposes of discussion, only the anisotropic diffusion (ADF) [54] and CNN are presented. Fig. 6 shows the results for the visual (EO) and infrared (MWIR) collected imagery and the ADF and CNN image fusion results for the first distance and a single view.

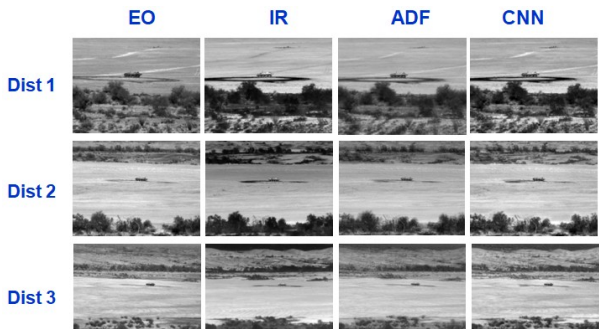

Fig. 6 Image Fusion Results

The aggregation of all the performance metrics for the two methods are shown in Table 4. The first observation was that the statistics were not consistent across all the image fusion methods and hence we chose two candidates with generally good performance: ADF and CNN. The second observation was that the ADF was typically the fastest (e.g., 0.054 s for distance 1) while the CNN was the slowest (e.g., 40.783 s for distance 1). The third observation is that most metrics are relatively the same for each method, so the image fusion choice is speed versus accuracy.

Table 4: Sample Statistics

| Method | MI | PSNR | GB | SSIM | SD | SF |
|---|---|---|---|---|---|---|
| | | | Distance 1 | | | |
| ADF | 2.10 | 62.03 | 0.39 | 1.54 | 52.74 | 6.17 |
| CNN | 1.99 | 61.14 | 0.47 | 1.44 | 60.33 | 10.15 |
| | | | Distance 2 | | | |
| ADF | 2.10 | 61.86 | 0.37 | 1.51 | 52.56 | 6.69 |
| CNN | 2.00 | 61.06 | 0.48 | 1.47 | 59.51 | 9.84 |
| | | | Distance 3 | | | |
| ADF | 2.02 | 62.59 | 0.38 | 1.53 | 47.59 | 6.92 |
| CNN | 1.93 | 61.63 | 0.46 | 1.48 | 51.73 | 9.27 |

\* higher better, MI-mutual information, PSNR - Peak signal-to-noise ratio, GB - Gradient-based fusion performance, SSIM - Structure Similarity, SD - Standard Deviation, SF - Spatial Frequency.

Fig. 7 presents a case of overall performance for the image fusion approaches. The analysis shows a clear performance bound over accuracy and distance (e.g., pixel resolution) that supports a suggested certifiable bound for the deployment of the AI image fusion methods. Thus, the certification of a "method" is difficult when the metrics (quantitative) are not consistent or the visual preference (quantitative) which is subjective. On the other hand, the comparison of image fusion (MWIR+VI) to non-image fusion (V only) could be a choice for certification of a distance less than 4 km as shown on the right in Fig. 7.

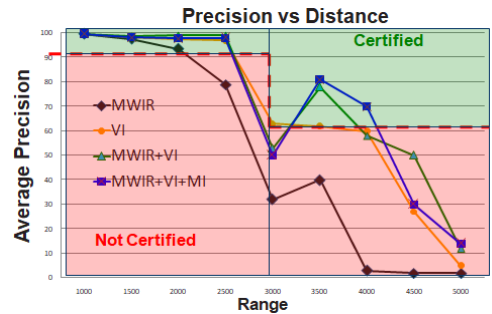

Fig. 7 Certifiable results for Image Fusion

## 7. Conclusions

The paper explored concepts in certification of AI/ML systems. Discussions from verifiable, explainable, and interpretable AI are useful to be combined with traditional certification methods of hardware, software, and users. The paper examined elements of human and machine certification, whether separately or together as a systems-level analysis. An example was presented from deep-learning based image fusion. Further research would examine the larger data set towards performance bounds associated with the processing with and without image fusion within an AI/ML architectural design.

**Acknowledgments**

The views and conclusions contained herein are those of the authors and should not be interpreted as necessarily representing the official policies or endorsements, either expressed or implied, of the author affiliations.